\documentclass[journal]{IEEEtran}
\usepackage{multirow}
\usepackage{amsmath}  
\usepackage{amssymb}  
\usepackage{graphicx} 
\usepackage{stfloats} 
\usepackage{booktabs}
\usepackage{caption} 
\usepackage{bbding}
\usepackage{makecell}
\usepackage{pifont}
\usepackage{array} 
\usepackage{lipsum} 
\usepackage{graphicx}
\usepackage{authblk}  
\usepackage{amssymb}
\usepackage[ruled,vlined]{algorithm2e}
\usepackage{multirow}
\usepackage{amsmath}
\usepackage{float}
\usepackage{enumitem}
\usepackage{amsmath}
\usepackage{authblk}
\usepackage{svg}

\ifCLASSINFOpdf
\else
\fi
\begin{document}
%
\title{DB-MSMUNet:Dual Branch Multi-scale Mamba UNet for Pancreatic CT Scans Segmentation}
\author{
  Qiu Guan\textsuperscript{1,\dag},\ 
  Zhiqiang Yang\textsuperscript{1,\dag},\ 
  Dezhang Ye\textsuperscript{1},\ 
  Yang Chen\textsuperscript{2,*},\ 
  Xinli Xu\textsuperscript{1},\ 
  Ying Tang\textsuperscript{1,*} \thanks{$^{\dag}$ These authors contributed equally to this work. $^{*}$ Corresponding authors: cheny@seu.edu.cn, ytang@zjut.edu.cn. This work is supported in part by the National Natural Science Foundation of China (62373324, U20A20171, 72192823, 61972355), the Key Project of Zhejiang Provincial Natural Science Foundation (LZ23F020010), and the Zhejiang Provincial “Jianbing Lingyan+X” Science and Technology Program (2025C01127).}
}
\affil{
  \textsuperscript{1} Zhejiang University of Technology, Hangzhou, China \\
  \textsuperscript{2} SouthEast University, Nanjing, China
}

\maketitle

\begin{abstract}
Accurate segmentation of the pancreas and its lesions in CT scans is crucial for the precise diagnosis and treatment of pancreatic cancer.
However, it remains a highly challenging task due to several factors such as low tissue contrast with surrounding organs, blurry anatomical boundaries, irregular organ shapes, and the small size of lesions. To tackle these issues, we propose DB-MSMUNet (Dual-Branch Multi-scale Mamba UNet), a novel encoder-decoder architecture designed specifically for robust pancreatic segmentation. The encoder is constructed using a Multi-scale Mamba Module (MSMM), which combines deformable convolutions and multi-scale state space modeling to enhance both global context modeling and local deformation adaptation. The network employs a dual-decoder design: the edge decoder introduces an Edge Enhancement Path (EEP) to explicitly capture boundary cues and refine fuzzy contours, while the area decoder incorporates a Multi-layer Decoder (MLD) to preserve fine-grained details and accurately reconstruct small lesions by leveraging multi-scale deep semantic features. Furthermore, Auxiliary Deep Supervision (ADS) heads are added at multiple scales to both decoders, providing more accurate gradient feedback and further enhancing the discriminative capability of multi-scale features. We conduct extensive experiments on three datasets: the NIH Pancreas dataset, the MSD dataset, and a clinical pancreatic tumor dataset provided by collaborating hospitals. DB-MSMUNet achieves Dice Similarity Coefficients of 89.47\%, 87.59\%, and 89.02\%, respectively, outperforming most existing state-of-the-art methods in terms of segmentation accuracy, edge preservation, and robustness across different datasets. These results demonstrate the effectiveness and generalizability of the proposed method for real-world pancreatic CT segmentation tasks.
\end{abstract}


%
\IEEEpeerreviewmaketitle

\section{Introduction}
%
%
%
%
\IEEEPARstart{P}{ancreatic} cancer presents significant diagnostic and therapeutic challenges, with a one-year survival rate of under 20\% and a five-year survival rate of less than 9\%~\cite{1}. CT scans are the principal modality for detecting pancreatic lesions, and timely surgical resection of tumors before they advance to pancreatic cancer is crucial for improving patient outcomes. Consequently, accurate segmentation of the pancreas and pancreatic tumors in CT images is crucial for effective clinical treatment. However, the pancreas constitutes a relatively small portion of abdominal CT images, often less than 1.5\% of a single slice. In these images, the pancreas and pancreatic tumors are closely adjacent to surrounding organs and blood vessels, sharing similar textures with neighboring tissues. This proximity and similarity in texture result in indistinct boundaries and low contrast, making accurate segmentation challenging. Consequently, a segmentation technique that can precisely delineate pancreatic lesions is essential for the effective treatment of pancreatic cancer.


The two most popular architectures in deep learning, namely convolutional neural networks (CNNs) and vision transformers (ViTs), are dominating the field of visual representation learning and has been widely applied to various medical image segmentation tasks~\cite{unet,attunet,transunet}. However, CNNs can effectively extract local features, they struggle to capture global context and long-term dependencies, leading to insufficient feature extraction. ViTs can effectively capture long-range dependencies, but their self-attention mechanism has high quadratic complexity in long sequence modeling, resulting in a heavy computational burden. In recent years, structured state-space models (SSMs)~\cite{zhu2024vision}, inspired by classical state-space models, have garnered widespread attention for their computational efficiency and excellent performance in modeling long-term dependencies. They have been widely applied to medical image segmentation tasks for various organs~\cite{ma2024u,ruan2024vm,xing2024segmamba}.
Nevertheless, transferring these Mamba-based networks to the task of pancreatic segmentation lacks dedicated optimization for pancreatic lesions, mainly due to the small size and deformation of the pancreas, thereby leaving considerable room for improvement.


Considering the aforementioned challenges, we incorporated deformable convolutions into the Mamba framework and proposed the Multi-scale Mamba Module. This block can dynamically adjust the regions of interest while effectively integrating global and local features, thereby addressing the deformation issues of the pancreas. Furthermore, to improve the model's ability to handle both fine-grained details and high-level semantics, we introduced a dual-decoder strategy. The dual-decoder strategy consists of two parallel decoders: the Edge Enhancement Path (EEP), which focuses on refining the pancreas edge details, and the Multi-layer Decoder (MLD), which targets small and subtle regions, especially in areas with low contrast or deformation. In addition, the Auxiliary Deep Supervision (ADS) heads facilitate more effective optimization of multi-scale feature representations in both decoders.

The main contributions of this work can be summarized as follows:
\begin{itemize}
\item[$\bullet$] We proposed the Multi-scale Mamba Module, which integrates deformable convolutions into the Mamba, addressing the deformation problem of the pancreas.
\item[$\bullet$] The Edge Enhancement Path aims to enhance the network's sensitivity to edge information by supervising the edge images of the pancreas.
\item[$\bullet$] We proposed a Multi-layer Decoder that upsamples the outputs of each layer of the backbone network, enabling the model to effectively reconstruct low-level features and address the issue of ignoring the target due to the small size of the pancreas.
\item[$\bullet$] Extensive experiments have demonstrated the effectiveness of the proposed DB-MSMUNet. In the NIH, MSD, and clinical pancreatic tumor datasets, our method has better performance.
\end{itemize}

\section{Related work}
\subsection{Pancreas segmentation}
Due to the pancreas’s similar texture to surrounding organs and low contrast with adjacent tissues, its boundaries are difficult to distinguish, posing significant challenges for accurate segmentation. Recently, several networks for pancreas segmentation have been proposed. 
Qiu et al.~\cite{qiu2023cmfcunet} introduced a cascaded segmentation network, referencing the structure of UNet3+ and incorporating a multi-scale feature calibration gate (MSCG) for feature fusion, achieving a 86.30 $\pm$ 4.03\% DSC on the NIH dataset. Wang et al.~\cite{wang2021pancreas} proposed a dual-input v-mesh network for pancreas segmentation, which generated edge-enhanced images using the GBVS algorithm to effectively solve the issue of blurred pancreatic edges. Additionally, deformable convolutions were employed to address the variability in pancreatic shape, ultimately achieving a DSC of 87.40 $\pm$ 6.80\% on the NIH dataset. However, these methods do not emphasize the integration of global and multi-scale features, and they overlook the problem of missing small targets caused by the encoder-decoder structure.

\subsection{Technology evolution based on Mamba}
As one of the most successful variants of SSM, Mamba has achieved modeling capabilities comparable to those of Transformers, while maintaining linear scalability with respect to sequence length. In recent years, it has also made significant progress in the field of medical imaging. Ruan et al.~\cite{ruan2024vm} proposed the first purely SSM-based medical image segmentation model VM-UNet, establishing a baseline for models solely based on SSM. Wang et al.~\cite{wang2024large} proposed a Large Kernel Mamba UNet (LKM-UNet), which enhances spatial modeling by assigning large receptive field kernels to SSM layers. They also introduced a bidirectional Mamba for position-aware sequence modeling, achieving an average DSC of 86.82\% on the Abdomen CT dataset. Xu et al.~\cite{xu2024hc} proposed a Hybrid Convolution Mamba model (HC Mamba) for medical image segmentation, combining multiple convolution techniques optimized for medical imaging to enhance the receptive field and reduce model parameters. It achieved a DSC of 88.18\% on the ISIC2017 dataset. However, these Mamba-based networks have performed well on most medical segmentation tasks, but they have not shown significant improvement in segmenting small organs like the pancreas, which are prone to deformation, have small volumes, and exhibit blurred edges.
\section{Method}
\subsection{Preliminaries}
Models based on SSM, namely the Structured State Space Sequence Model (S4) and Mamba, originate from continuous systems that map one-dimensional sequences \( x(t) \to y(t) \) through hidden states \( h(t) \in \mathbb{R}^N \). This process can be represented by the following linear ordinary differential equation:
\begin{equation}
h'(t) = A h(t) + B x(t),  y(t) = C h(t).
\end{equation}  
where \( A \in \mathbb{R}^{N \times N} \) is a state matrix and \( B, C \in \mathbb{R}^N \) are projection parameters.
S4 and Mamba represent the discrete counterparts of the previously mentioned continuous system, incorporating a timescale parameter \( \Delta \) to convert the continuous parameters \( A \) and \( B \) into their discrete counterparts $\bar{A}$, $\bar{B}$. Generally, the zero-order hold (ZOH) method is utilized for discretization, which can be described as follows:
\begin{equation}
\bar{A} = \exp(\Delta A), \\
\bar{B} = (\Delta A)^{-1} \left(\exp(\Delta A) - I\right) \cdot \Delta B.
\end{equation}
Following discretization, the discrete form of Equation (1) is defined as:
\begin{equation}
h'(t) = \bar{A} h(t) + \bar{B} x(t), y(t) = C h(t).
\end{equation}
Subsequently, the output is obtained using a global convolution, which is defined as:
\begin{equation}
K = \left(C \bar{B}, \; C \overline{AB}, \; C \bar{A}^{L-1} \bar{B} \right), y = x \ast \bar{K}
\end{equation}
where \( L \) represents the length of the input sequence \( x \), and \( K \in \mathbb{R}^L \) denotes a structured convolutional kernel.
\subsection{Overall Framework of DB-MSMUNet}

\begin{figure*}[htbp]
    \centering
    \includegraphics[width=0.7\textwidth]{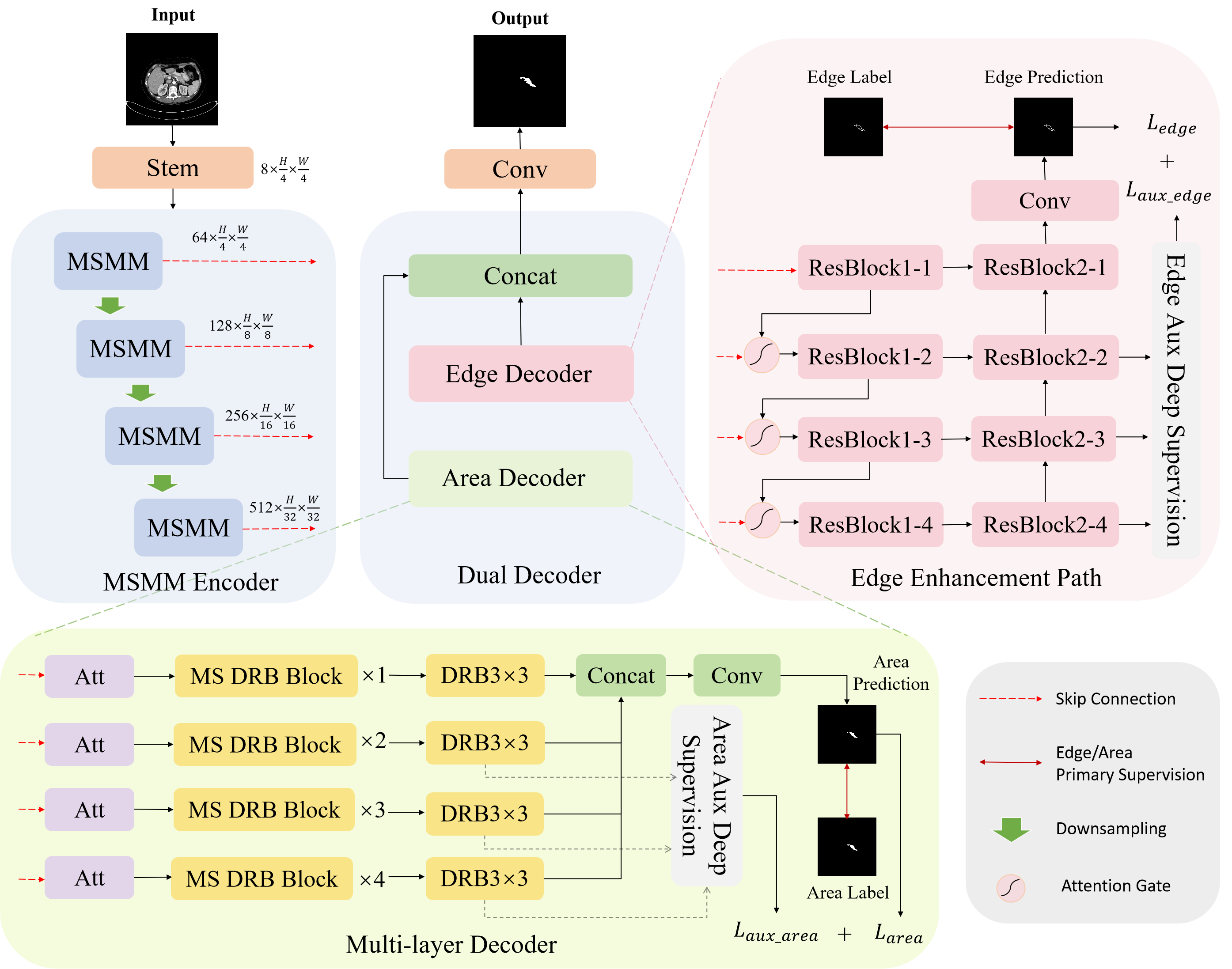} 
    \caption{The overall framework of DB-MSMUNet.}
    \label{DB-MSMUNet}
\end{figure*}

Fig.~\ref{DB-MSMUNet} illustrates the overall architecture of DB-MSMUNet, a dual-branch multi-scale Mamba network model proposed in this paper. The input image is first processed by a Stem block to extract basic features and reduce computational cost. 
Next, the Multi-scale Mamba Module (MSMM) serves as the network backbone, extracting multi-scale features through receptive fields of different sizes to capture both local and global context, which is vital for representing complex pancreatic structures. After feature extraction, two parallel decoders are employed: the Edge Enhancement Path (EEP), which refines lesion boundaries, and the Multi-layer Decoder (MLD), which reconstructs fine details and alleviates semantic gaps between feature levels. The final layers of both decoders generate edge and area losses, while Auxiliary Deep Supervision (ADS) provides multi-scale guidance. All losses are combined to obtain the final total loss.

\subsection{Multi-scale Mamba Module}
\begin{figure}[h] 
    \centering
    \includegraphics[width=0.38\textwidth]{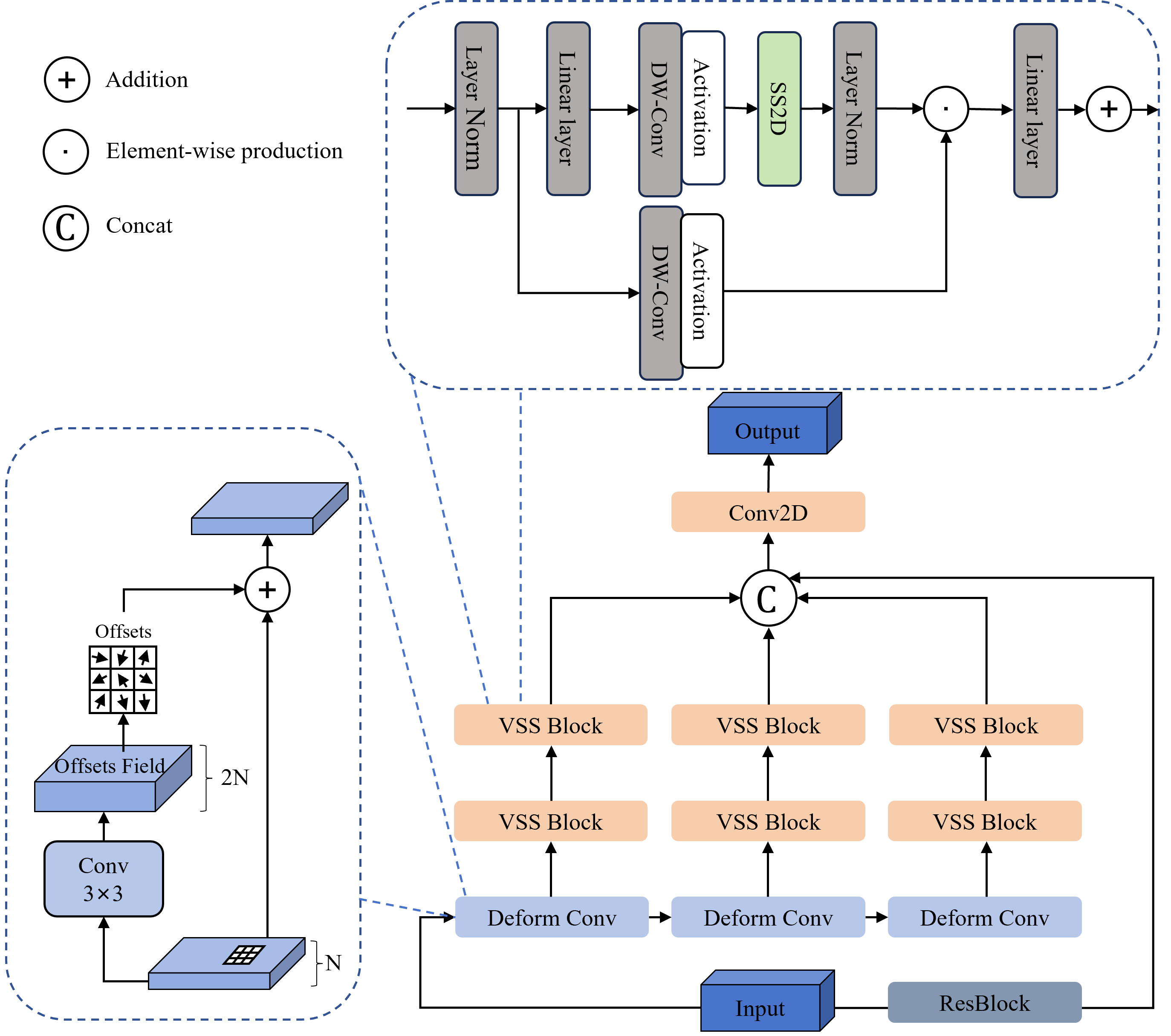} 
    \caption{Structure diagram of MSMM.} 
    \label{MSMM} 
\end{figure}
In pancreastic CT scans, both coarse-grained and fine-grained features are crucial, and network design needs to consider both larger-scale positional and shape features and smaller-scale texture features. However, the current Transformer and Mamba architectures cannot capture different fine-grained features simultaneously.

To address this issue, we proposes a multi-scale Mamba module that simulates different receptive fields by changing the size of convolutional kernels to obtain features at different scales from the input image. In this module, we employ three sequential 3$\times$3 deformable convolutions to represent three distinct receptive fields, with each convolution capturing receptive fields of 3$\times$3, 5$\times$5, and 7$\times$7, respectively. Unlike regular convolutions, deformable convolutions add a deformable offset field that contains a learnable offset for each position in the feature map. The added deformable offset field enhances the network's ability to extract features, enabling it to adaptively match the shape of the pancreas. By integrating deformable convolutions into Mamba, this module can flexibly capture the morphological differences of the pancreas, thereby achieving high-precision segmentation.

The overall structure of the proposed Multi-scale Mamba Module is shown in Fig.~\ref{MSMM}. After generating feature maps with three different receptive fields using deformable convolutions, we send them to the different two-layer Mamba module. Finally, the feature maps from multiple branches are concatenated together and input to the next layer.

Each layer computation of MSMM can be represented as
\begin{equation}
G_{i,j}=Mamba(F_{k \times k}(X_i)),j=0,1,2.k=3,5,7.
\end{equation}
\begin{equation}
L_{i}=Res(X_i)
\end{equation}
\begin{equation}
X_{i+1}=Concat(L_{i},G_{i,0},G_{i,1}...G_{i,j})
\end{equation}

In the equation, \(X_i \in \mathbb{R}^{\frac{H'}{2^i} \times \frac{W'}{2^i} \times 2^iC'}\) represents the output of the \(i\)-th layer, where \(H'\), \(W'\), and \(C'\) respectively represent the three dimensions of the feature map after undergoing the Stem processing.$F_{k \times k}$ represents the feature processed by a ${k \times k}$ deformable convolution, the value of \(k\) is taken as 3, 5, and 7. $Mamba$ represents the feature processed by Mamba Encoding, \(G_{i,j}\) represents the j-th global feature of the i-th layer. For example, \(G_{i,0}\) represents the first global feature obtained from \(X_i\) through $F_{3 \times 3}$ convolution and Mamba Encoding. $Res$ stands for ResBlock, and $L_i$ represents the local feature of the i-th layer.



\subsection{Edge Enhancement Path}

\begin{figure}[h] 
    \centering
    \includegraphics[width=\columnwidth]{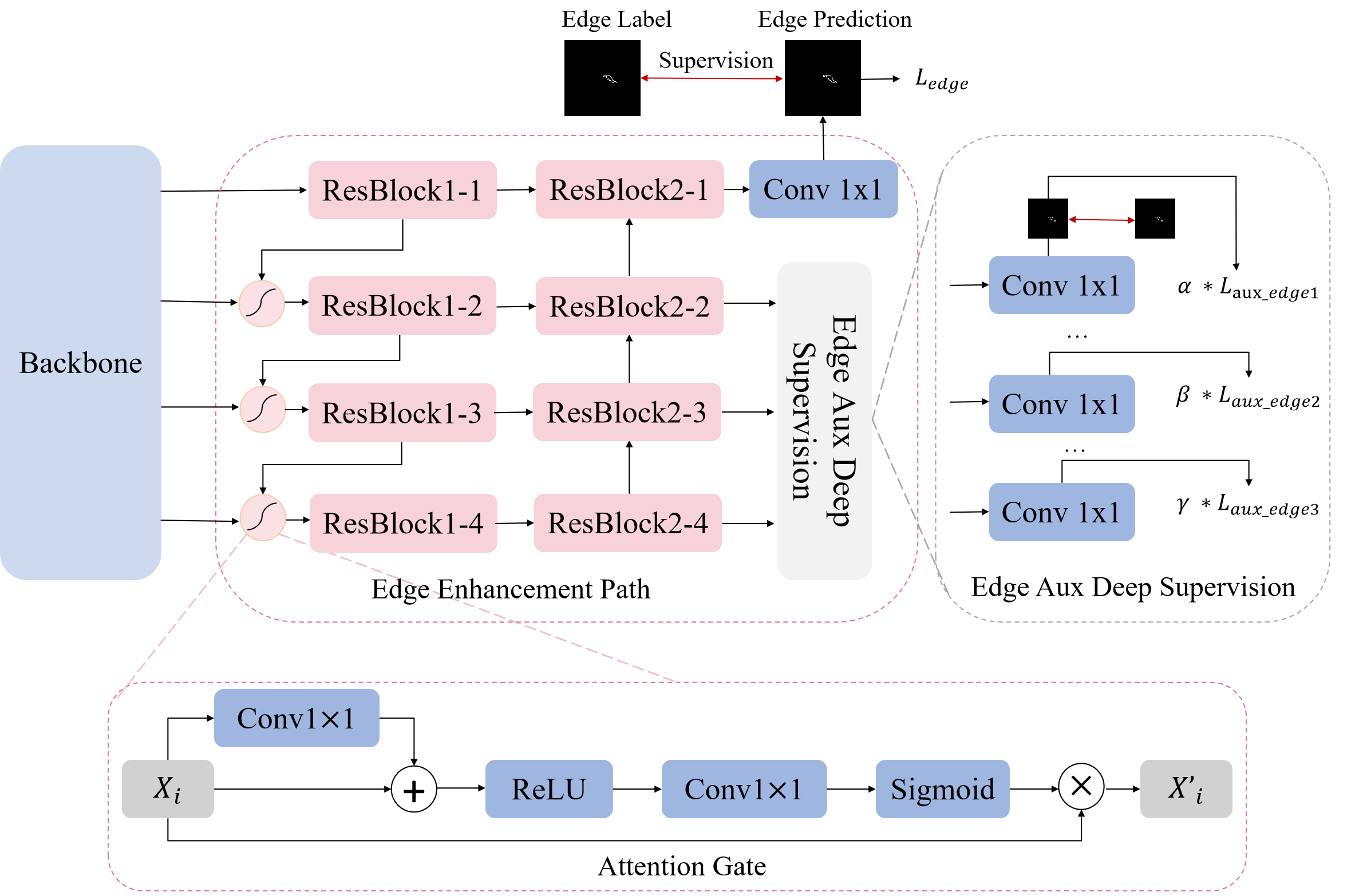} 
    \caption{Overall architecture of Edge Enhancement Path.} 
    \label{EEP} 
\end{figure}

Poor boundary contour segmentation poses a significant challenge in pancreatic segmentation. Traditional U-Net models, through successive downsampling, often lose edge details, resulting in discontinuous boundaries in the segmentation output, which can impact clinical diagnosis. Therefore, explicit modeling of the edges is necessary to enhance the boundary response. To address this issue, we propose the Edge Enhancement Path (EEP) to strengthen the backbone’s learning of pancreatic boundary contour information, as shown in Fig.~\ref{EEP}. 

Specifically, for each layer output ${X_i}$ of the backbone, we incorporate residual blocks to process and refine the relevant boundary and shape information, and apply an Attention Gate to ensure that the edge information is exclusively focused on processing boundary-related details. The extraction of edge feature information at each layer is as follows:
\begin{equation}
A_i = \sigma \left( \text{Conv}_{(1 \times 1)} \left( \text{ReLU} \left( \text{Conv}_{(1 \times 1)} (X_i) + X_i \right) \right) \right)
\end{equation}
\begin{equation}
X_i' = A_i \otimes X_i
\end{equation}
\begin{equation}
X_i = \text{ResBlock}_{(1-i)} (X_i')
\end{equation}
\begin{equation}
X_i' = \text{ResBlock}_{(2-i)} (X_{(i+1)}') + X_i
\end{equation}
where \(\sigma\) denotes the sigmoid activation function, \(\otimes\) represents the element-wise product, \(\text{ResBlock}\) denotes the residual convolution block, \(F_1'\) represents the edge prediction result, and \(i \in \{1, 2, 3, 4\}\) represents the different layers of the network.

For the edge auxiliary deep supervision, in order to balance the contributions of the final output head and the auxiliary layers, we introduce weighting factors $\alpha$, $\beta$, and $\gamma$ as hyperparameters. By default, their values are set to 0.6, 0.3, and 0.1, respectively. The final edge auxiliary loss is computed as:
\begin{equation}
\mathcal{L}_{\text{aux\_edge}} = \alpha \cdot \mathcal{L}_{\text{aux\_edge1}} + \beta \cdot \mathcal{L}_{\text{aux\_edge2}} + \gamma \cdot \mathcal{L}_{\text{aux\_edge3}}
\end{equation}

\subsection{Multi-layer Decoder}
The pancreas occupies a small portion of abdominal CT images, and its tail has a slender shape. After multiple downsampling operations in deep networks, the resolution of the pancreas gradually decreases, leading to a loss of small target features. Additionally, traditional U-shaped decoder structures, due to the skip connections that simply add feature maps of the same size from different levels, cannot effectively integrate the different semantics of upper and lower layers. This can even introduce noise or other interference, reducing segmentation accuracy in small target areas. Moreover, a single upsampling path is insufficient to collect enough effective multi-scale information, which may ultimately result in a blurred or broken boundary in the overall representation of the target, making it difficult to segment the complex morphology of the pancreas.
\begin{figure}[htbp]
    \centering
    \includegraphics[width=0.45\textwidth]{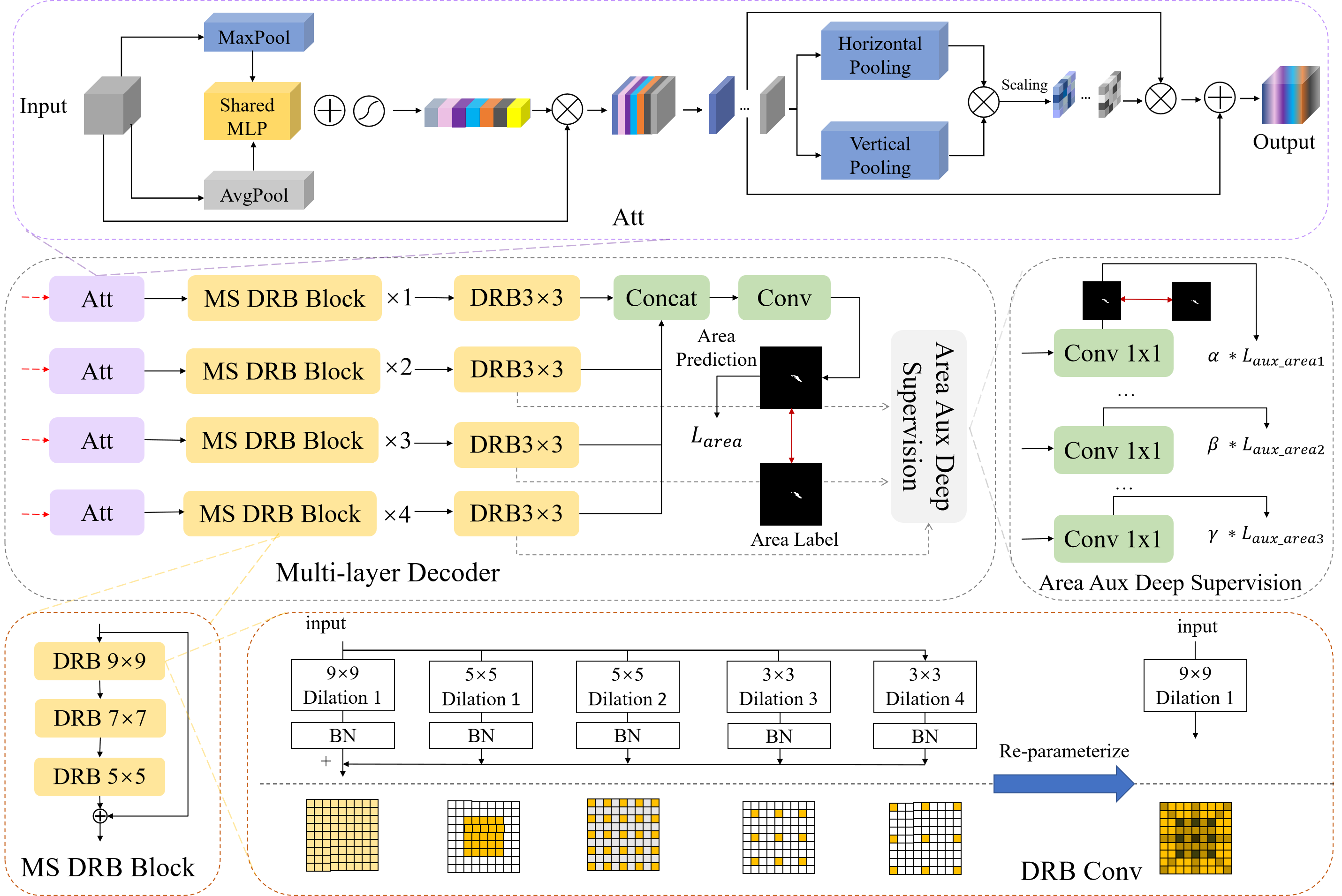} 
    \caption{Multi-layer decoder structure diagram.}
    \label{MLD}
\end{figure}
To address these challenges, we propose an E-shaped Multi-layer Decoder (MLD), as shown in Fig.~\ref{MLD}. First, the MLD receives output features from the encoder at different levels and processes them through a dual-attention module to adaptively adjust the feature map weights. The features then pass through multiple dilated reparameterized convolution modules~\cite{ding2024unireplknet} for upsampling, which helps recover more detailed information. Finally, the processed feature maps are passed into the decoder's output layer, resulting in precise segmentation. The process of the MLD is as follows:

\begin{equation}
D_i={MS DRB}({DRB}_{3\times 3}^{(i)}(E_i))(i=1,2,3,4)
\end{equation}
\begin{equation}
O=Conv(Concat(D_1,D_2,D_3,D_4))
\end{equation}

\( E_i \) represents the feature map \( X_i \) after being processed by the dual-attention mechanism \( \text{Att} \). DRB represents the Dilated Re-parameterization convolutional, where different kernel sizes, such as \(9 \times 9\), \(7 \times 7\), and \(5 \times 5\), are used to capture multi-scale features and enhance the feature extraction process. The MSDRB refers to a feature extraction module composed of three sequential DRB convolutional layers. The detailed parameter design can be found in the original paper~\cite{ding2024unireplknet}. \(O\) denotes the output of the Multi-layer Decoder (MLD). And the auxiliary loss computation for the area auxiliary deep supervision heads is similar to that of the edge auxiliary loss, using the same set of weighting factors.

\section{Experimental results}
\subsection{Datasets}
We conducted pancreas and tumor segmentation experiments on the NIH~\cite{26}, MSD2018~\cite{27}, and clinical datasets. The NIH dataset contains 82 contrast-enhanced abdominal CT scans, divided into 61 for training and 21 for validation. The MSD dataset includes 281 scans, with 211 used for training and 70 for testing. For consistency, pancreas and tumor labels were merged into a single category. Additionally, 89 contrast-enhanced CT scans with pancreatic tumor labels were collected from the First Affiliated Hospital of Zhejiang University School of Medicine for clinical evaluation. Since our method is designed for 2D images, all volumetric data were sliced along the horizontal plane, resulting in 7,309, 9,073, and 1,476 2D images, respectively.

\subsection{Implementation details}
We implemented our method based on the PyTorch platform on the Ubuntu system equipped with an NVIDIA GeForce RTX 4090 graphics card of 24 GB memory.

For model training, we chose AdamW as the optimizer of our network, the initial learning rate was set to 0.0005. In addition, the learning rate is adjusted using the Cosine Annealing strategy, with a maximum period of 32 epochs, and updates occurring after each epoch. We set the epoch number to 300 and the batch size to 14. 

In our experiments, we use two types of loss: the area loss and the edge loss. The area loss is defined as Dice loss, while the edge loss is given by:
\begin{equation}
\begin{aligned}
\mathcal{L}_{\text{edge}} = & \sum_{(x,y)} \left( w_0 \cdot E_{(x,y)} \cdot \log(P_{(x,y)}) \right) \\
& + \sum_{(x,y)} \left( w_1 \cdot (1 - E_{(x,y)}) \cdot \log(1 - P_{(x,y)}) \right)
\end{aligned}
\end{equation}
\begin{equation}
w_0 = \frac{\sum_{(x,y)} E_{(x,y)}}{W \cdot H}, \quad w_1 = 1 - w_0
\end{equation}
Here, \(E_{(x,y)}\) represents the edge label at position \((x, y)\), \(P_{(x,y)}\) represents the predicted edge probability at position \((x, y)\), and \(w_0\) and \(w_1\) represent the weights for labels 0 and 1, respectively. \(W\) and \(H\) denote the width and height of the label image. The total loss is then defined as:

\begin{equation}
\mathcal{L}_{\text{total}} = \mathcal{L}_{\text{area}} +  \mathcal{L}_{\text{aux\_area}} +\mathcal{L}_{\text{edge}} + \mathcal{L}_{\text{aux\_edge}} 
\end{equation}

For data preprocessing, we adjusted the window level and width to capture grayscale values accurately. The NIH and MSD datasets were clipped to \([-100,+240]\) HU, and the clinical dataset to \([-100,+140]\) HU, then normalized to \([0,255]\). Data augmentation included random flipping, $90^\circ$ rotation, Gaussian noise, contrast adjustment, Gaussian smoothing, and histogram shifting. Edge labels were generated using the Canny operator, which extracted region boundaries to produce binary edge maps.


\begin{table*}[t]
    \centering
    \renewcommand\arraystretch{1.3}
    \caption{Comparison of Segmentation Results with Other SOTA Network Models on Three Datasets.}
    \label{SOTA_comparison_table}
    \setlength{\tabcolsep}{2mm}  
    \begin{tabular}{l>{\centering\arraybackslash}p{1.2cm}>{\centering\arraybackslash}p{1.2cm}>{\centering\arraybackslash}p{1.2cm}>{\centering\arraybackslash}p{1.2cm}>{\centering\arraybackslash}p{1.2cm}>{\centering\arraybackslash}p{1.2cm}>{\centering\arraybackslash}p{1.2cm}>{\centering\arraybackslash}p{1.2cm}>{\centering\arraybackslash}p{1.2cm}}
    \hline
    \multirow{2}{*}{\textbf{Network Model}} & \multicolumn{3}{c}{\textbf{NIH}} & \multicolumn{3}{c}{\textbf{MSD}} & \multicolumn{3}{c}{\textbf{Clinical}} \\ \cline{2-10} 
    & DSC(\%) & P(\%) & R(\%) & DSC(\%) & P(\%) & R(\%) & DSC(\%) & P(\%) & R(\%) \\ \hline
    UNet~\cite{unet} & 80.14 & 83.64 & 78.64 & 81.46 & 83.64 & 78.64 & 77.03 & 83.33 & 82.24 \\
    nnU-Net~\cite{nnuet} & 85.34 & 85.68 & 88.32 & 85.38 & 87.12 & 88.07 & 85.91 & 87.06 & 89.11 \\
    TransUNet~\cite{transunet} & 83.18 & 84.84 & 89.15 & 82.58 & 85.69 & 87.01 & 80.82 & 91.05 & 85.30 \\ 
    SwinUNETR~\cite{swin} & 83.64 & 84.08 & 85.14 & 83.26 & 84.79 & 87.98 & 85.43 & 90.65 & 86.57 \\ 
    VM-UNet~\cite{ruan2024vm} & 82.71 & 84.28 & 89.52 & 84.27 & 85.87 & 86.45 & 83.87 & 91.52 & 85.64 \\ 
    U-Mamba~\cite{ma2024u} & 85.31 & 87.10 & 90.43 & 84.31 & 86.17 & 85.79 & 84.14 & 90.71 & 84.76 \\ 
    SliceMamba~\cite{slicemamba} & 87.09 & 88.01 & 90.13 & 86.01 & 88.25 & 86.98 & 85.34 & 90.99 & 85.53 \\ \hline
    Ours & \textbf{89.47} & \textbf{90.24} & \textbf{92.04} & \textbf{87.59} & \textbf{88.98} & \textbf{89.02} & \textbf{89.02} & \textbf{92.34} & \textbf{91.72} \\ \hline

    \end{tabular}
\end{table*}

\subsection{Segmentation results on three datasets}
To evaluate the effectiveness of the proposed method, we compared it with other competitive approaches on the NIH, MSD, and clinical datasets. All experiments were performed using four-fold cross-validation, and the reported results represent the average performance across all folds.

\begin{figure*}[htbp]
    \centering
    \includegraphics[width=0.7\textwidth]{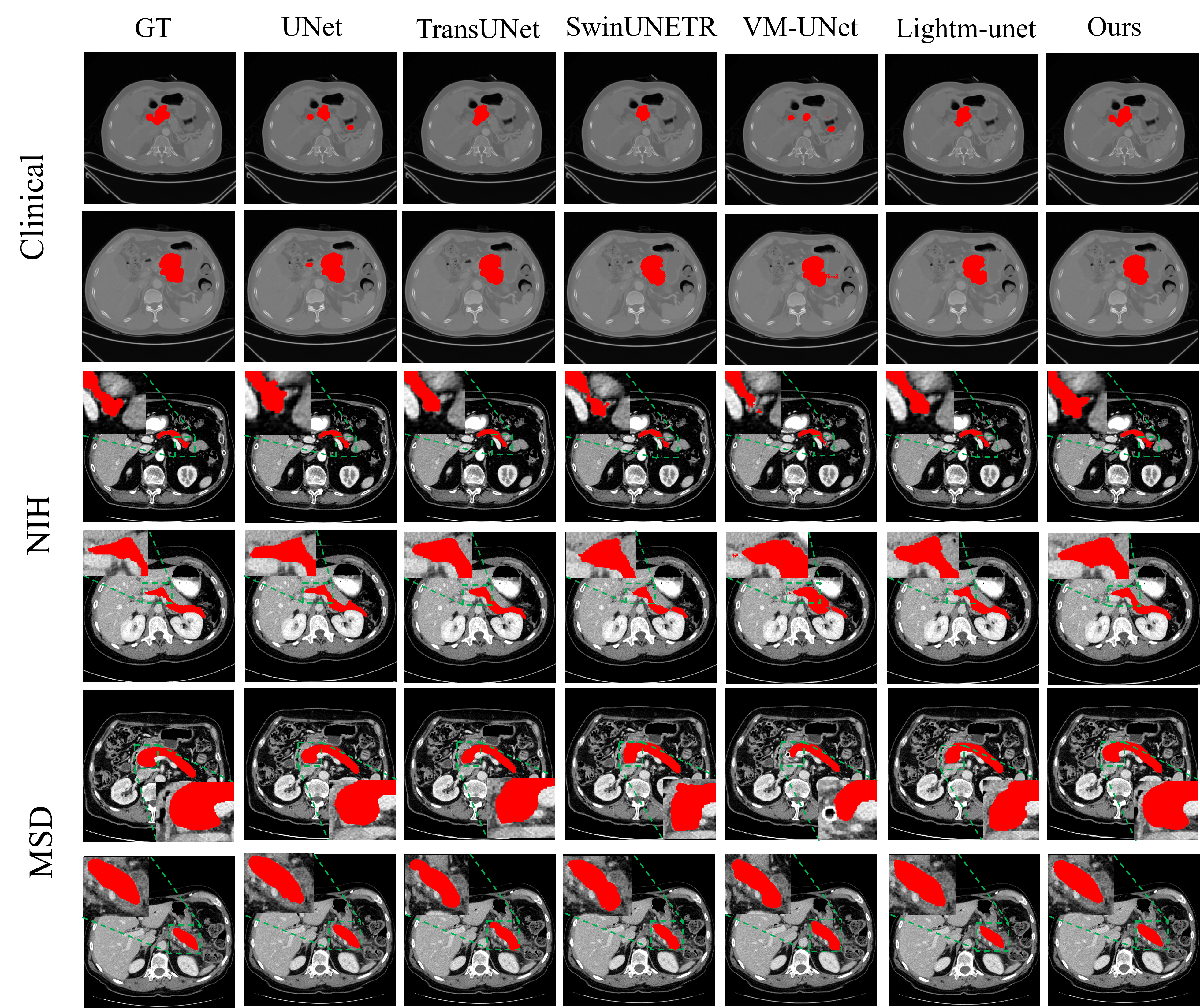} 
    \caption{The performance comparison across the three datasets.}
    \label{result}
\end{figure*}

From the results in Table~\ref{SOTA_comparison_table}, it can be seen that the proposed method outperforms traditional segmentation models like UNet and nnU-Net, as well as current state-of-the-art Transformer-based models such as TransUNet and SwinUNETR, and Mamba-based models like VM-UNet, U-Mamba and SliceMamba in terms of DSC, Precision, and Recall. Compared to Transformer-based models, our method shows improvements across all three datasets, with a notable 3.59\% increase in DSC, 1.69\% improvement in Precision, and 5.15\% gain in Recall on the Clinical dataset. When compared to newer Mamba-based models such as SliceMamba, our method also demonstrates superior performance, with a 2.38\% increase in DSC, 2.23\% improvement in Precision, and 1.91\% in Recall on the NIH dataset. These results highlight the effectiveness of the edge-enhanced decoder and multi-layer decoder in improving the model’s ability to detect small targets and edge structures, leading to more accurate and finer segmentation. As shown in the visual results in Fig.~\ref{result}, our method reduces false positives and improves boundary recognition, especially in complex and low-contrast regions, showing higher segmentation precision and robustness.


\subsection{Ablation experiments}
To validate the effectiveness of the proposed Multi-scale Mamba Module (MSMM), Edge Enhancement Path (EEP) and Multi-layer Decoder (MLD) in this study, we conducted ablation experiments on the NIH dataset. As shown in Table~\ref{tab:example}, we removed each of the proposed three innovative modules from the network individually and measured the segmentation DSC metric of the remaining network. 
\begin{figure*}[htbp]
    \centering
    \includegraphics[width=0.7\textwidth]{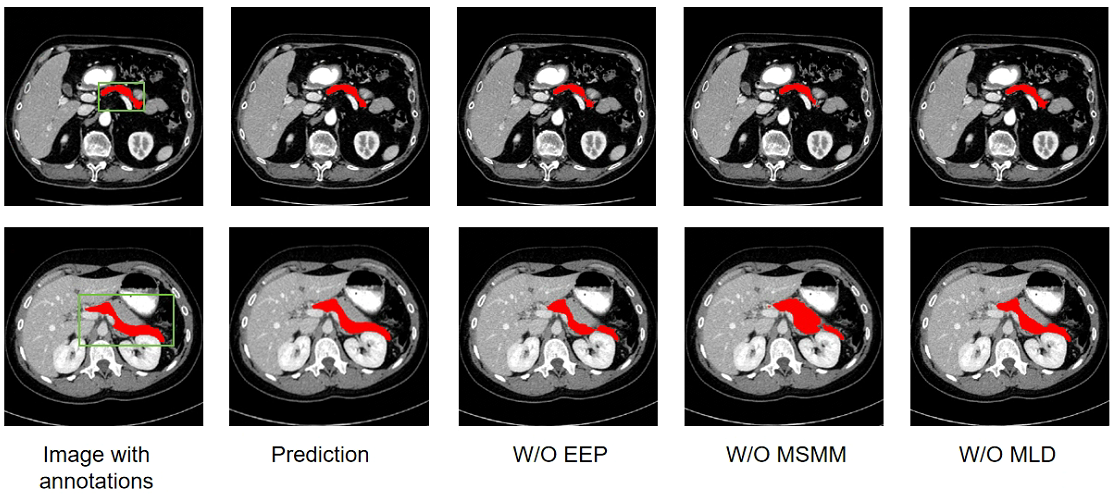} 
    \caption{Demonstration of the ablation study results on the NIH dataset segmentation. The leftmost column shows the overlay of the original image and ground truth. 'W/O' indicates the absence of the module.}
    \label{ablation}
\end{figure*}

\begin{table}[h!]
\centering    
\caption{Ablation experiments. "-" for MSMM means we use a single path Mamba instead. "-" for MLD means we use a normal UNet Decoder instead.}
\renewcommand\arraystretch{1.3}
\begin{tabular}{ccccc}

\hline
\quad MSMM \quad & \quad EEP \quad & \quad MLD \quad & \quad ADS \quad & \quad DSC(\%)\quad \\
\hline
\checkmark & - & - & - &86.23 \\
- & \checkmark & \checkmark & - &86.19 \\
\checkmark & - & \checkmark & - &86.68 \\
\checkmark & \checkmark & - & - &87.95 \\
\checkmark & \checkmark & \checkmark & - &88.99 \\
\checkmark & \checkmark & \checkmark & \checkmark &\textbf{89.47} \\
\hline
\end{tabular}
\label{tab:example}
\end{table}

\begin{table}[h!]
\centering    
\caption{Comparative experiments of different segmentation methods using various backbone.}
\renewcommand\arraystretch{1.3}
\begin{tabular}{cccc}
\hline
Method &  Backbone &  Param  & DSC(\%) \\   
\hline

\multirow{2}{*} {U-Mamba\_Bot} & U-Mamba-encoder &63M &85.31 \\  
& MSMM-encoder  &45M&86.84\\
\hline
 \multirow{4}{*}{DB-MSMUNet} & nnU-Net-encoder &46M &86.20 \\
& UNETR-encoder~\cite{unetr}  &73M&85.75\\
& U-Mamba-encoder &62M&86.97\\
& MSMM-encoder &\textbf{44M}&\textbf{89.47}\\
\hline
\end{tabular}
\label{tab:example1}
\end{table}

In Table~\ref{tab:example}, we observed that the absence of the MSMM led to the most significant decrease in segmentation results, reaching 3.28\%, demonstrating the crucial role of our proposed MSMM in segmentation. Subsequently, the removal of the EEP resulted in a 2.79\% decrease in segmentation performance. Additionally, the 1.52\% improvement in segmentation results indicates the usefulness of MLD. Finally, with the introduction of ADS, an additional improvement of 0.48\% is achieved. To more intuitively demonstrate the impact of the proposed innovative modules on the network, we visualized the network segmentation performance after dropping a single submodule in Fig.~\ref{ablation}. The information presented in Fig.~\ref{ablation} is as follows: First, the network achieves relatively complete pancreatic segmentation using the MSMM, addressing the issue of pancreatic deformation to a certain extent. Second, EEP further refines the segmentation of pancreatic edge contours by supervising the backbone network. Finally, MLD is employed to improve the reconstruction of small pancreatic lesions, ultimately leading to accurate pancreatic segmentation. The visualization of segmentation results further confirms the effectiveness of each innovative module, which is consistent with the data presented in Table~\ref{tab:example}. 

To validate the advantages of our proposed MSMM-Encoder as a backbone network, we conducted comparative experiments using U-Mamba Bot and DB-MSMUNet as baseline models, as shown in Table~\ref{tab:example1}. The goal is to analyze the benefits of our model in terms of both parameter efficiency and segmentation performance.
For the U-Mamba Bot model, our MSMM-Encoder reduces the parameter count by 18 million compared to the original U-Mamba Encoder, while achieving a 1.53\% improvement in performance.
In the case of DB-MSMUNet, the MSMM-Encoder achieves the best performance with only 44 million parameters, outperforming CNN-based, Transformer-based, and Mamba-based backbones by 3.27\%, 3.72\%, and 2.50\%, respectively.


\section{Conclusion}
This paper proposes a Multi-scale Mamba UNet for pancreatic segmentation, effectively addressing challenges such as pancreatic deformation, small organ size, and low contrast that lead to blurred boundaries. Specifically, a module combining deformable convolution with Mamba captures broader contextual information and adapts to shape variations, improving segmentation accuracy. The Edge Enhancement Path (EEP) focuses on refining pancreatic boundary contours, while the Multi-layer Decoder (MLD) preserves shallow semantic details and reconstructs fine features. Experiments on the NIH, MSD, and clinical datasets show that the proposed MSMUNet achieves competitive results, surpassing most SOTA models.

Despite its strong performance, some limitations remain. Although the MSMM-Encoder greatly reduces parameters, the dual-decoder still relies on CNN modules, resulting in higher computational cost. Replacing the decoder with a Mamba-based design led to performance degradation. Future work will aim to further simplify and optimize the decoder while maintaining high segmentation accuracy.
\ifCLASSOPTIONcaptionsoff
  \newpage
\fi



%
\bibliographystyle{splncs04}
\bibliography{ref}

\begin{thebibliography}{10}
\providecommand{\url}[1]{\texttt{#1}}
\providecommand{\urlprefix}{URL }
\providecommand{\doi}[1]{https://doi.org/#1}

\bibitem{transunet}
Chen, J., Lu, Y., Yu, Q., Luo, X., Adeli, E., Wang, Y., Lu, L., Yuille, A.L., Zhou, Y.: Transunet: Transformers make strong encoders for medical image segmentation. arXiv preprint arXiv:2102.04306  (2021)

\bibitem{ding2024unireplknet}
Ding, X., Zhang, Y., Ge, Y., Zhao, S., Song, L., Yue, X., Shan, Y.: Unireplknet: A universal perception large-kernel convnet for audio video point cloud time-series and image recognition. In: Proceedings of the IEEE/CVF Conference on Computer Vision and Pattern Recognition. pp. 5513--5524 (2024)

\bibitem{slicemamba}
Fan, C., Yu, H., Huang, Y., Wang, L., Yang, Z., Jia, X.: Slicemamba with neural architecture search for medical image segmentation. IEEE Journal of Biomedical and Health Informatics  (2025)

\bibitem{swin}
Hatamizadeh, A., Nath, V., Tang, Y., Yang, D., Roth, H.R., Xu, D.: Swin unetr: Swin transformers for semantic segmentation of brain tumors in mri images. In: Brainlesion: Glioma, Multiple Sclerosis, Stroke and Traumatic Brain Injuries: 7th International Workshop, BrainLes 2021, Held in Conjunction with MICCAI 2021, Virtual Event, September 27, 2021, Revised Selected Papers, Part I. pp. 272--284. Springer (2022)

\bibitem{unetr}
Hatamizadeh, A., Tang, Y., Nath, V., Yang, D., Myronenko, A., Landman, B., Roth, H.R., Xu, D.: Unetr: Transformers for 3d medical image segmentation. In: Proceedings of the IEEE/CVF winter conference on applications of computer vision. pp. 574--584 (2022)

\bibitem{nnuet}
Isensee, F., Petersen, J., Klein, A., Zimmerer, D., Jaeger, P.F., Kohl, S., Wasserthal, J., Koehler, G., Norajitra, T., Wirkert, S., et~al.: nnu-net: Self-adapting framework for u-net-based medical image segmentation. arXiv preprint arXiv:1809.10486  (2018)

\bibitem{ma2024u}
Ma, J., Li, F., Wang, B.: U-mamba: Enhancing long-range dependency for biomedical image segmentation. arXiv preprint arXiv:2401.04722  (2024)

\bibitem{attunet}
Oktay, O., Schlemper, J., Folgoc, L.L., Lee, M., Heinrich, M., Misawa, K., Mori, K., McDonagh, S., Hammerla, N.Y., Kainz, B., et~al.: Attention u-net: Learning where to look for the pancreas. arXiv preprint arXiv:1804.03999  (2018)

\bibitem{qiu2023cmfcunet}
Qiu, C., Song, Y., Liu, Z., Yin, J., Han, K., Liu, Y.: Cmfcunet: cascaded multi-scale feature calibration unet for pancreas segmentation. Multimedia Systems  \textbf{29}(2),  871--886 (2023)

\bibitem{unet}
Ronneberger, O., Fischer, P., Brox, T.: U-net: Convolutional networks for biomedical image segmentation. In: Medical Image Computing and Computer-Assisted Intervention--MICCAI 2015: 18th International Conference, Munich, Germany, October 5-9, 2015, Proceedings, Part III 18. pp. 234--241. Springer (2015)

\bibitem{26}
Roth, H.R., Lu, L., Farag, A., Shin, H.C., Liu, J., Turkbey, E.B., Summers, R.M.: Deeporgan: Multi-level deep convolutional networks for automated pancreas segmentation. In: Medical Image Computing and Computer-Assisted Intervention--MICCAI 2015: 18th International Conference, Munich, Germany, October 5-9, 2015, Proceedings, Part I 18. pp. 556--564. Springer (2015)

\bibitem{ruan2024vm}
Ruan, J., Xiang, S.: Vm-unet: Vision mamba unet for medical image segmentation. arXiv preprint arXiv:2402.02491  (2024)

\bibitem{1}
Siegel, R.L., Miller, K.D., Jemal, A.: Cancer statistics, 2019. CA: a cancer journal for clinicians  \textbf{69}(1),  7--34 (2019)

\bibitem{27}
Simpson, A.L., Antonelli, M., Bakas, S., Bilello, M., Farahani, K., Van~Ginneken, B., Kopp-Schneider, A., Landman, B.A., Litjens, G., Menze, B., et~al.: A large annotated medical image dataset for the development and evaluation of segmentation algorithms. arXiv preprint arXiv:1902.09063  (2019)

\bibitem{wang2024large}
Wang, J., Chen, J., Chen, D., Wu, J.: Large window-based mamba unet for medical image segmentation: Beyond convolution and self-attention. arXiv preprint arXiv:2403.07332  (2024)

\bibitem{wang2021pancreas}
Wang, Y., Gong, G., Kong, D., Li, Q., Dai, J., Zhang, H., Qu, J., Liu, X., Xue, J.: Pancreas segmentation using a dual-input v-mesh network. Medical Image Analysis  \textbf{69},  101958 (2021)

\bibitem{xing2024segmamba}
Xing, Z., Ye, T., Yang, Y., Liu, G., Zhu, L.: Segmamba: Long-range sequential modeling mamba for 3d medical image segmentation. arXiv preprint arXiv:2401.13560  (2024)

\bibitem{xu2024hc}
Xu, J.: Hc-mamba: Vision mamba with hybrid convolutional techniques for medical image segmentation. arXiv preprint arXiv:2405.05007  (2024)

\bibitem{zhu2024vision}
Zhu, L., Liao, B., Zhang, Q., Wang, X., Liu, W., Wang, X.: Vision mamba: Efficient visual representation learning with bidirectional state space model. arXiv preprint arXiv:2401.09417  (2024)

\end{thebibliography}



%








\end{document}